\documentclass[11pt]{article}

\usepackage[final]{acl}
\usepackage{latexsym}
\usepackage{microtype}

\usepackage[T1]{fontenc}
\usepackage[utf8]{inputenc}
\usepackage[english]{babel}
\usepackage{times}

\usepackage{graphicx}
\usepackage{amsmath}
\usepackage{amssymb}
\usepackage{booktabs}
\usepackage{simpleicons}
\usepackage{fontawesome5}
\usepackage{multirow}
\usepackage{colortbl}
\usepackage{tabularx}
\usepackage{enumitem}
\usepackage{xcolor}
\usepackage{placeins}
\usepackage{adjustbox}

\definecolor{worse}{HTML}{E8963C}
\definecolor{better}{HTML}{3AA88B}

\title{Alignment without Transfer: Tokenization Impacts Cross-lingual Representations during Multilingual Language Model Pretraining}

\author{
Jonathan von Rad \quad Pontus Stenetorp \quad Yao Lu\thanks{Corresponding author.} \\
\textnormal{University College London, Centre for Artificial Intelligence} \\
{\footnotesize\texttt{jonathan.rad.25@ucl.ac.uk}} \\
{\footnotesize\texttt{p.stenetorp@cs.ucl.ac.uk \quad yao.lu@cs.ucl.ac.uk }}
}

\begin{document}
\maketitle

\newcommand{\BTS}{
\begin{table}[t]
\centering
\small
\renewcommand{\arraystretch}{1.35}
\setlength{\tabcolsep}{3pt}
\begin{adjustbox}{max width=\columnwidth}
\begin{tabular}{ll ccc ccc}
\toprule
 & & \multicolumn{3}{c}{into partner language} & \multicolumn{3}{c}{into English} \\
\cmidrule(lr){3-5} \cmidrule(lr){6-8}
pair & script & starved & fair & $\Delta$ & starved & fair & $\Delta$ \\
\midrule
en-de & same  & $+.000$ & $-.005$ & $-.005$ & $-.030$ & $-.019$ & $+.011$ \\
en-fr & same  & $+.015$ & $+.008$ & $-.007$ & $-.018$ & $-.020$ & $-.003$ \\
en-ar & cross & $-.009$ & $-.013$ & $-.004$ & $-.039$ & $-.032$ & $+.007$ \\
en-zh & cross & $-.013$ & $-.017$ & $-.004$ & $-.024$ & $-.026$ & $-.002$ \\
\midrule
\multicolumn{2}{l}{mean gap (same $-$ cross)} & $+.018$ & $+.016$ & $-.002$ & $+.007$ & $+.009$ & $+.002$ \\
\multicolumn{2}{l}{worst same $-$ best cross} & $+.009$ & $+.008$ & $-.001$ & $-.006$ & $+.006$ & $+.012$ \\
\bottomrule
\end{tabular}
\end{adjustbox}
\caption{$\mathrm{BTS}^{\text{loss}}$ at matched own-language exposure
($d = 11.1$B per language, holdout BPB) under both tokenizers, for transfer
into the partner language and into English. The mean gap is the average of
the two same-script pairs minus the average of the two cross-script pairs;
the conservative margin is the \emph{worst} same-script pair minus the
\emph{best} cross-script pair. $\Delta$ is fair minus starved.}
\label{tab:bts-both-directions}
\end{table}
}

\newcommand{\TokenizerEffects}{
\begin{table}[h]
\centering
\small
\resizebox{\columnwidth}{!}{%
\renewcommand{\arraystretch}{1.5}
\begin{tabular}{l rr rr}
\toprule
 & \multicolumn{2}{c}{Into partner language} & \multicolumn{2}{c}{Into English} \\
\cmidrule(lr){2-3}\cmidrule(lr){4-5}
Tokenizer & BTS$^{\text{loss}}$ & Effect [95\% CI] & BTS$^{\text{loss}}$ & Effect [95\% CI] \\
\midrule
419 languages & $-0.19$ & $+0.36$ [$-0.10$, $+0.83$] & $-2.76$ & $-0.12$ [$-0.50$, $+0.26$] \\
20 languages  & $-0.69$ & $-0.14$ [$-0.60$, $+0.33$] & $-3.08$ & $-0.44$ [$-0.82$, $-0.06$] \\
5 languages   & $-0.68$ & $-0.13$ [$-0.59$, $+0.34$] & $-2.43$ & $+0.21$ [$-0.17$, $+0.58$] \\
2 languages   & $-0.65$ & $-0.10$ [$-0.57$, $+0.37$] & $-2.29$ & $+0.35$ [$-0.03$, $+0.73$] \\
\midrule
Grand mean    & $-0.55$ & & $-2.64$ & \\
$F(3,9)$      & \multicolumn{2}{c}{$1.05$ ($p = 0.42$)} & \multicolumn{2}{c}{$3.33$ ($p = 0.07$)} \\
\bottomrule
\end{tabular}%
}
\caption{
Tokenizer effects on cross-lingual transfer (BTS$^{\text{loss}}$, pp) at $\approx$11B own-language tokens. Effects are deviations from the four-pair grand mean, with 95\% CIs from a pair\,+\,tokenizer model using the interaction residual (9 df) as error.
}
\label{tab:tok-effects}
\end{table}
}

\newcommand{\HFmodel}[2]{%
  \href{https://huggingface.co/jvonrad/alignment-without-transfer/tree/main/model-checkpoints/#1}{#2}%
}

\newcommand{\ModelList}{
\begin{table}[t]
\centering
\small
\begin{tabular}{@{}lcccc@{}}
\toprule
& \multicolumn{4}{c}{Tokenizer languages} \\
\cmidrule(lr){2-5}
Model & 419 & 20 & 5 & 2 \\
\midrule
\multicolumn{5}{@{}l}{\textbf{Monolingual (23 models)}} \\
\addlinespace[2pt]
EN
 & \HFmodel{monolingual/en/419-languages/30b-final}{30B}
 & \HFmodel{monolingual/en/20-languages/12b}{12B}
 & \HFmodel{monolingual/en/5-languages/30b-final}{30B}
 & \shortstack{
   \HFmodel{monolingual/en/2-languages-en-de/12b}{DE},
   \HFmodel{monolingual/en/2-languages-en-fr/12b}{FR} \\
   \HFmodel{monolingual/en/2-languages-en-ar/12b}{AR},
   \HFmodel{monolingual/en/2-languages-en-zh/12b}{ZH}
   } \\
DE
 & \HFmodel{monolingual/de/419-languages/16b}{16B}
 & \HFmodel{monolingual/de/20-languages/12b}{12B}
 & \HFmodel{monolingual/de/5-languages/30b-final}{30B}
 & \HFmodel{monolingual/de/2-languages/12b}{12B} \\
FR
 & \HFmodel{monolingual/fr/419-languages/30b-final}{30B}
 & \HFmodel{monolingual/fr/20-languages/12b}{12B}
 & \HFmodel{monolingual/fr/5-languages/30b-final}{30B}
 & \HFmodel{monolingual/fr/2-languages/12b}{12B} \\
AR
 & \HFmodel{monolingual/ar/419-languages/30b-final}{30B}
 & \HFmodel{monolingual/ar/20-languages/12b}{12B}
 & \HFmodel{monolingual/ar/5-languages/30b-final}{30B}
 & \HFmodel{monolingual/ar/2-languages/12b}{12B} \\
ZH
 & \HFmodel{monolingual/zh/419-languages/15b}{15B}
 & \HFmodel{monolingual/zh/20-languages/12b}{12B}
 & \HFmodel{monolingual/zh/5-languages/15b}{15B}
 & \HFmodel{monolingual/zh/2-languages/12b}{12B} \\
\midrule
\multicolumn{5}{@{}l}{\textbf{Bilingual (16 models)}} \\
\addlinespace[2pt]
EN--DE
 & \HFmodel{bilingual/en-de/419-languages/30b-final}{30B}
 & \HFmodel{bilingual/en-de/20-languages/30b-final}{30B}
 & \HFmodel{bilingual/en-de/5-languages/30b-final}{30B}
 & \HFmodel{bilingual/en-de/2-languages/30b-final}{30B} \\
EN--FR
 & \HFmodel{bilingual/en-fr/419-languages/30b-final}{30B}
 & \HFmodel{bilingual/en-fr/20-languages/30b-final}{30B}
 & \HFmodel{bilingual/en-fr/5-languages/30b-final}{30B}
 & \HFmodel{bilingual/en-fr/2-languages/30b-final}{30B} \\
EN--AR
 & \HFmodel{bilingual/en-ar/419-languages/30b-final}{30B}
 & \HFmodel{bilingual/en-ar/20-languages/30b-final}{30B}
 & \HFmodel{bilingual/en-ar/5-languages/30b-final}{30B}
 & \HFmodel{bilingual/en-ar/2-languages/30b-final}{30B} \\
EN--ZH
 & \HFmodel{bilingual/en-zh/419-languages/30b-final}{30B}
 & \HFmodel{bilingual/en-zh/20-languages/30b-final}{30B}
 & \HFmodel{bilingual/en-zh/5-languages/30b-final}{30B}
 & \HFmodel{bilingual/en-zh/2-languages/30b-final}{30B} \\
\bottomrule
\end{tabular}%

\caption{Overview of all 39 pretrained models across
four tokenizer conditions. Each entry links to the
latest available Hugging Face checkpoint and shows
its training budget in billions of tokens (B).
For English under the 2-language condition,
links indicate the partner-language tokenizer. Intermediate checkpoints can be found under the same HF repo.}
\label{tab:model-checkpoints}
\end{table}

}

\begin{abstract}
Tokenization represents one of the most crucial steps in mulilingual NLP pipelines. Yet, tokenizers routinely under-allocate vocabulary to low-resource languages. While improved allocation leads to better performance, it remains poorly understood how tokenization shapes cross-lingual representations and transfer during pretraining. To address this, we pretrain 39 mono- and bilingual 1B decoder-only models on FineWeb2-HQ across English, German, French, Arabic and Chinese, varying only the tokenizer. Our results demonstrate that improved vocabulary allocation (i) substantially raises cross-lingual representation alignment in early-to-mid layers, an effect that prior work measuring only the output layer could not observe, (ii) increases cross-lingual parametric knowledge sharing in feed-forward neurons, supported by correlational and causal evidence, and (iii) an \emph{alignment-transfer dissociation}, where cross-lingual transfer remains essentially unchanged despite higher alignment. Moreover, transfer is significantly lower for cross-script bilinguals even under fair vocabulary allocation, indicating that the script barrier is not merely a tokenizer artifact. All of our checkpoints and code are publicly available. \footnote{
\simpleicon{huggingface}\ \href{https://huggingface.co/jvonrad/xscript-eval}{\texttt{jvonrad/xscript-eval} \\
\faGithub\ \href{https://github.com/jvonrad/XScript-Pretraining}{\texttt{jvonrad/XScript-Pretraining}} \\}
}
\end{abstract}

\begin{figure}[t]
\centering
\includegraphics[width=\columnwidth]{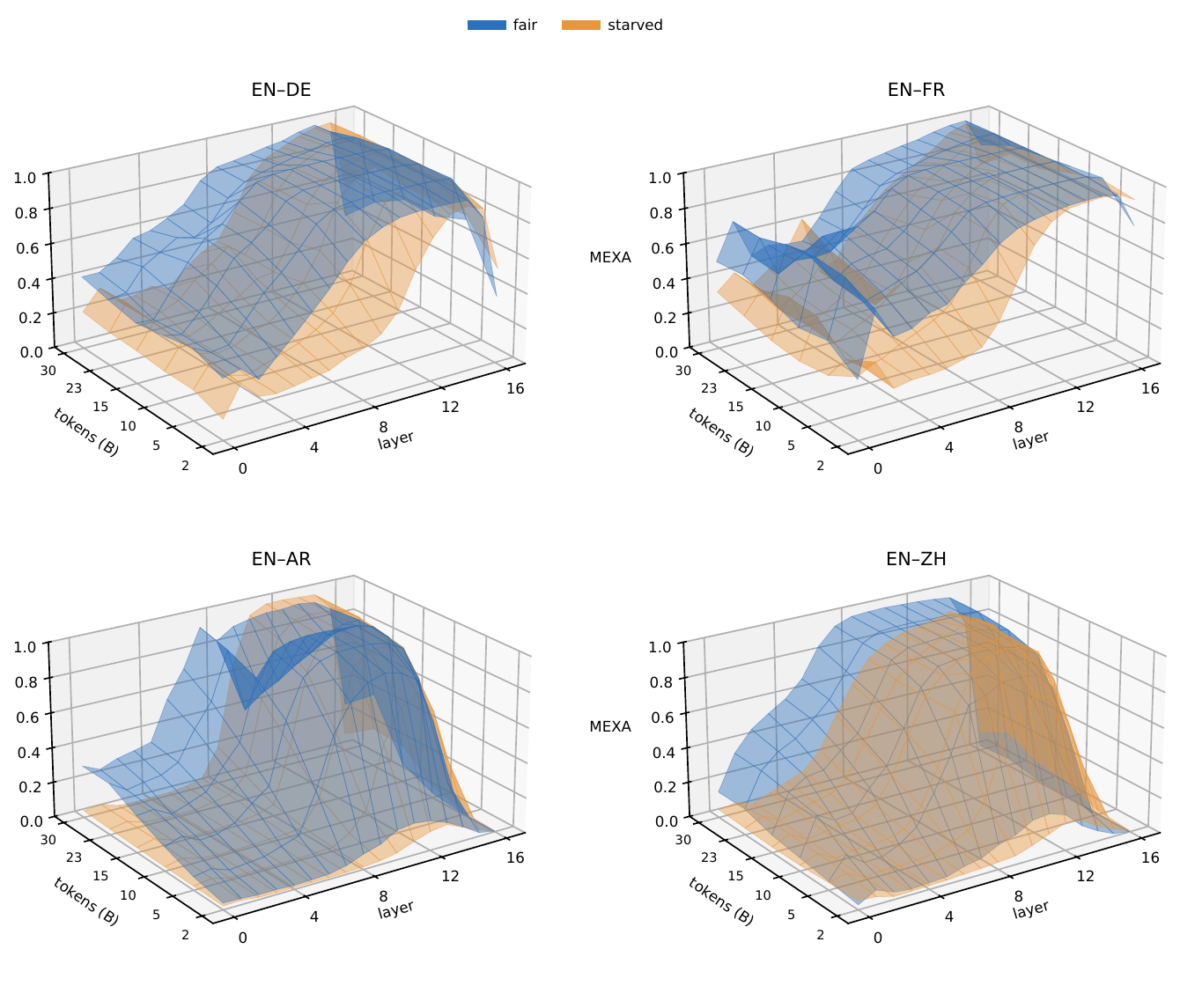}
\caption{Cross-lingual representation alignment for a \emph{fair} and \emph{starved} tokenizer over layer and training tokens for each bilingual on its own language pair measured by MEXA \cite{kargaran-etal-2025-mexa}.}
\label{fig:mexa-3d}
\end{figure}

\section{Introduction}
\label{sec:intro}

Large language models (LLMs) exhibit strong capabilities in English but struggle in low-resource languages, particularly those with different scripts \cite{xuan-etal-2025-mmlu, han-etal-2026-mubench}. Since most languages lack sufficient training data to develop these capabilities independently, they rely heavily on cross-lingual transfer from high-resource languages \cite{joshi-etal-2020-state, hu-etal-2025-quantifying}. Another important factor in multilingual pretraining is \emph{tokenization}: how text is segmented into subword units. Tokenizers optimized for high-resource languages often allocate insufficient vocabulary to other languages, reducing tokenization efficiency and downstream performance \cite{arnett-bergen-2025-language, ali-etal-2024-tokenizer}. 
Yet how tokenization shapes cross-lingual alignment and transfer during pretraining remains poorly understood. Alignment encompasses the similarity of representations across languages, the consistency of model outputs, and the sharing of knowledge within model parameters. While these properties may facilitate the reuse of learned capabilities across languages, whether stronger alignment actually improves cross-lingual transfer remains debated \cite{hua-etal-2024-mothello, gaschi-etal-2023-exploring}. In particular, it is unclear whether improving vocabulary allocation strengthens alignment and, if so, whether these improvements translate into better cross-lingual transfer.

To investigate this, we pretrain 39 monolingual and bilingual 1B parameter decoder-only models across five languages and four tokenizers, holding the vocabulary size fixed at 64k while varying vocabulary allocation through the tokenizer training corpus. We make \textbf{four main contributions}:

\begin{enumerate}

\item We show that improved vocabulary allocation substantially increases representation alignment across languages in early-to-mid layers.

\item We reveal an \emph{alignment-transfer dissociation} during decoder-only pretraining, where improved vocabulary allocation increases alignment, but leaves cross-lingual transfer unchanged. 

\item We further show that better tokenization increases cross-lingual parametric knowledge sharing in MLP neurons, supported by correlational and causal evidence.

\item We conduct a controlled pretraining study of 39 monolingual and bilingual 1B models across five languages and four tokenizers, and will publicly release all model checkpoints and code upon publication.

\end{enumerate}


\section{Related Work}
\label{sec:background}

\paragraph{Multilingual Tokenization.}
Multilingual tokenizers often disproportionately allocate vocabulary to high-resource languages, resulting in less efficient segmentation and reduced downstream performance for underrepresented languages \cite{ali-etal-2024-tokenizer, ahia-etal-2023-languages, arnett-bergen-2025-language}. While vocabulary size influences multilingual performance, \citet{limisiewicz-etal-2023-tokenization} show that language-specific vocabulary allocation, rather than vocabulary size alone, is a key determinant of tokenizer quality. These inequalities are further amplified by differences in UTF-8 encoding across scripts, motivating byte-premium adjustments \cite{arnett-etal-2024-bit} and more equitable tokenizer training objectives \cite{foroutan-etal-2026-parity}. However, while previous work has extensively studied the effects of tokenization on performance and efficiency, its influence on cross-lingual representations, parametric knowledge sharing, and transfer during pretraining remains poorly understood.

\paragraph{Cross-lingual Transfer and the Script Barrier.}
Multilingual pretraining can benefit low-resource languages while negatively affecting high-resource ones, a phenomenon associated with the \emph{curse of multilinguality} \cite{conneau-etal-2020-unsupervised, chang-etal-2024-multilinguality, malkin-etal-2022-balanced}. Recent work measures cross-lingual transfer directly during pretraining by comparing monolingual and multilingual models under matched target-language exposure \cite{chang-etal-2024-multilinguality, longpre2026atlas}. This is done by, for instance, comparing validation loss of a monolingual model trained on 15B French (target-language) tokens, with a bilingual model trained on 15B French (target-language) tokens and 15B English (source-language) tokens. These studies identify both script disparity and source-language resource availability as predictors of cross-lingual transfer. The reduction in transfer associated with script disparity is commonly referred to as the \emph{script barrier} \cite{longpre2026atlas, xhelili-etal-2024-breaking}. However, these studies pay little attention to how their tokenizers allocate vocabulary across languages. When a tokenizer splits a source language’s text into more tokens, the model sees less of that text under a fixed token budget, even when target-language exposure is matched. Some of the apparent transfer differences may therefore reflect differences in tokenization efficiency.  To overcome the script barrier, transliteration has been explored as one approach to increase transfer across languages of different scripts \cite{xhelili-etal-2024-breaking, moosa-etal-2023-transliteration}. In addition, data quality systematically varies with resource-language \cite{kreutzer-etal-2022-quality, messmer-etal-2025-enhancing}. We build on the pretraining-based transfer framework of \citet{longpre2026atlas}, isolating the effects of vocabulary allocation under controlled data quality and model capacity.

\paragraph{Cross-lingual Representation Alignment.}
Cross-lingual representation alignment measures the similarity of hidden representations for semantically equivalent inputs across languages \cite{conneau-etal-2020-emerging, muller-etal-2021-first}. Recent methods such as MEXA \cite{kargaran-etal-2025-mexa} quantify alignment through translation retrieval, accounting for limitations of raw cosine similarity in anisotropic representation spaces \cite{ethayarajh-2019-contextual, haemmerl-etal-2023-exploring}. Previous studies have linked vocabulary overlap and tokenizer design to alignment, but have focused on encoders \cite{ kautsar2026paralleltokenizersrethinkingencoder} or only the output layer \cite{limisiewicz-etal-2023-tokenization}, reporting no effect of vocabulary allocation on sentence retrieval (which is used to measure representation alignment), rather than alignment dynamics across all network layers throughout decoder-only pretraining. Moreover, whether stronger alignment facilitates cross-lingual transfer remains contested. While \citet{gaschi-etal-2023-exploring} report a positive correlation, \citet{hua-etal-2024-mothello} demonstrate that alignment can emerge without transfer, and \citet{veitsman2026bettercrosslingualalignmentfails} find that improved alignment does not necessarily facilitate task transfer. Our work investigates this relationship by tracking alignment across layers and training stages while experimentally varying vocabulary allocation.

\paragraph{Cross-lingual Knowledge Sharing.}
\citet{ifergan-etal-2025-beneath} show that knowledge can remain encoded in language-specific parameters, particularly across scripts. Mechanistic interpretability studies have identified feed-forward networks (FFNs) as important components of factual knowledge storage \cite{geva-etal-2023-dissecting, locating2022meng}. Knowledge neurons can be identified through attribution methods such as integrated gradients \cite{sundararajan2017axiomatic, dai-etal-2022-knowledge}, which have been extended to identify language-independent knowledge neurons \cite{chen2023journeycenterknowledgeneurons}. Related work investigates language-specific neurons \cite{tang-etal-2024-language} and cross-lingual knowledge conductivity \cite{gao-etal-2024-multilingual}. Yet how vocabulary allocation affects parametric knowledge sharing remains unclear. We address this through neuron overlap measurements and a tokenizer-invariant causal knowledge transfer analysis, examining whether improved vocabulary allocation promotes functional knowledge sharing across languages.

\section{Method}
\label{sec:method}

\subsection{Experimental Design}
\label{sec:design}

To isolate the effects of vocabulary allocation on cross-lingual alignment and transfer, we pretrain 39 monolingual and bilingual models across four tokenizer conditions (seven tokenizers total). Following \citet{longpre2026atlas}, we contrast a replication of their \emph{starved} tokenizer covering 419 languages with a \emph{fair} tokenizer trained on our five study languages. We introduce two additional conditions covering 20 and 2 languages to assess whether alignment changes systematically with vocabulary allocation.

\paragraph{Languages and Data.}
We study English (\textsc{en}), German (\textsc{de}), French (\textsc{fr}), Arabic (\textsc{ar}), and Mandarin Chinese (\textsc{zh}), spanning different language families and scripts. English serves as the anchor, yielding two same-script pairs (\textsc{en--de}, \textsc{en--fr}) and two cross-script pairs (\textsc{en--ar}, \textsc{en--zh}). To control for data quality, we use FineWeb2-HQ \cite{messmer-etal-2025-enhancing} for non-English languages and FineWeb-HQ for English, both curated using the same filtering pipeline.

\begin{figure}[t]
\centering
\includegraphics[width=\linewidth]{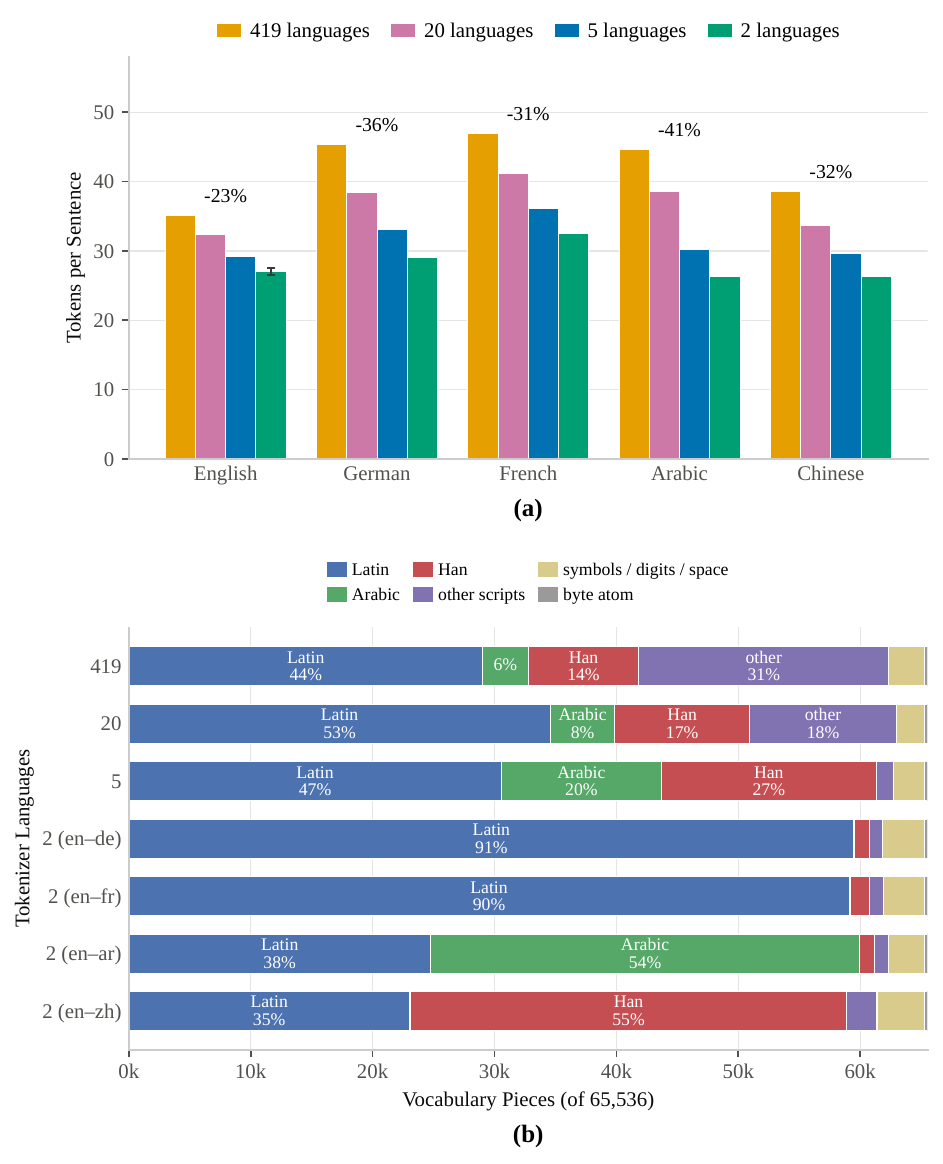}
\caption{Four tokenizer conditions trained on different numbers of languages. (a)~Fertility measured in tokens per sentence on FLORES+. (b)~Vocabulary share by script.}
\label{fig:tokenizer-comparison}
\end{figure}

\paragraph{Tokenizers.}
All tokenizers use SentencePiece Unigram \cite{kudo-richardson-2018-sentencepiece} with a fixed 64k vocabulary, matching \citet{longpre2026atlas} and keeping embedding and unembedding dimensions constant. Figure~\ref{fig:tokenizer-comparison} compares fertility and vocabulary allocation across conditions. Our \emph{starved} tokenizer replicates the 419-language coverage and sampling procedure of \citet{longpre2026atlas}, using temperature $T=100$ and FineWeb2 data. The \emph{fair} tokenizer instead allocates its vocabulary across our five languages, using byte-premium ratios \cite{arnett-etal-2024-bit} to approximately equalize semantic content during sampling. We additionally train a 20-language tokenizer and four pair-specific bilingual tokenizers in order to have a wider spectrum of vocabulary allocation. Appendix~\ref{app:tokenizer-replication} demonstrates the faithfulness of our replication against the original tokenizer from \citet{longpre2026atlas}.

\subsection{Models and Training}
\label{sec:models}

We pretrain 39 LLaMA-style \cite{touvron2023llama2openfoundation} decoder-only models at the 1B-parameter scale. Under each tokenizer, we train five monolingual and four English-centered bilingual models for evaluating performance, transfer, and alignment. Under the four pair-specific 2-language tokenizers, we train eight monolingual models (four English and four partner-language baselines) and four bilingual models. For a list of all trained models with open-source links, refer to Appendix \ref{app:model-list}, Table \ref{tab:model-checkpoints}.

All models share 16 Transformer layers, hidden size 2048, 16 attention heads, RMSNorm, SwiGLU, and rotary position embeddings. Training uses AdamW \cite{loshchilov2018decoupled} in bf16 for up to 30B tokens, with a warmup-stable-decay schedule \cite{scaling2022hagele} comprising 1B warmup, 23B stable, and 6B decay tokens. Bilingual models sample equally from both languages. Training required 4,330 GPU-h on Nvidia GH200 GPUs and 2,930 Chip-h on Trainium 2.48x Accelerators. Further details appear in Appendix~\ref{app:setup}.

\subsection{Evaluation}
\label{sec:evaluation}

\begin{figure}[h]
    \centering
    \includegraphics[width=\linewidth]{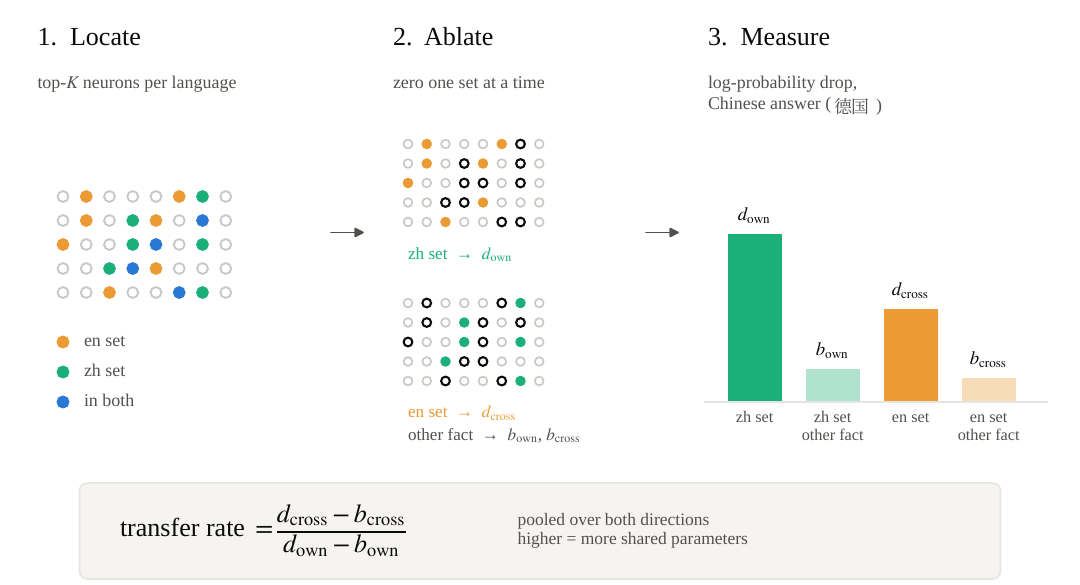}
    \caption{Causal test of parametric sharing for ``\emph{Which country in Europe has the biggest population? Answer: Germany}''. Subtracting a different-fact baseline $b$ controls for neurons supporting factual recall generally, while normalization enables comparison across tokenizers.}
    \label{fig:causal-sharing}
\end{figure}

\paragraph{Cross-lingual Knowledge Sharing.}
For facts correctly answered in both languages on the PolyFact dataset \cite{vonrad2026polyfactcomparingconsistencydrivenposttraining}, we identify the top $K$ knowledge neurons in feed-forward networks using integrated gradients \cite{dai-etal-2022-knowledge, chen2023journeycenterknowledgeneurons}, following \citet{chen2023journeycenterknowledgeneurons}. We then separately ablate the top 100 neurons identified in source language $s$ and target language $t$ (Figure~\ref{fig:causal-sharing}). We measure the resulting drops in the gold answer's log-likelihood in $t$, denoted $d_{\mathrm{cross}}$ and $d_{\mathrm{own}}$. Subtracting the corresponding different-fact ablation effects, $b_{\mathrm{cross}}$ and $b_{\mathrm{own}}$, controls for general factual-recall contributions. We define the causal ablation transfer rate as

\begin{equation}
\mathrm{CTR}_{s\to t}
= \frac{d_{\mathrm{cross}} - b_{\mathrm{cross}}}
       {d_{\mathrm{own}} - b_{\mathrm{own}}}.
\label{eq:causal-transfer-rate}
\end{equation}

Higher CTR indicates greater cross-lingual sharing of fact-specific neurons. We compare neuron overlap and CTR across the fair and starved tokenizers.

\paragraph{Bilingual Transfer.}
Since tokenizers produce different numbers of tokens per text, we measure validation loss in bits-per-byte (BPB) \cite{gao2020pile800gbdatasetdiverse} rather than per-token cross-entropy. We quantify transfer from source language $s$ to target language $t$ by comparing the target-language validation loss of monolingual and bilingual models:

\begin{equation}
\mathrm{BTS}^{\text{loss}}_{s \to t}(d)
= \frac{L^{\text{mono}}_t(d)-L^{\text{bi}}_t(d)}
{L^{\text{mono}}_t(d)},
\label{eq:bts-loss}
\end{equation}

where positive values indicate beneficial transfer and negative values indicate interference. Comparisons are restricted to the stable WSD phase ($1$--$24$B tokens), ensuring identical learning-rate states across checkpoints. 
Unlike the BTS formulation of \citet{longpre2026atlas}, which measures the training tokens required to reach a reference loss, our formulation compares losses directly at matched target-language exposure, which avoids undefined transfer scores when the reference loss is never reached.

\paragraph{Representation Alignment.}
We measure cross-lingual alignment using MEXA \cite{kargaran-etal-2025-mexa} on FLORES+ \cite{gordeev-etal-2024-flores}. At each layer $l$, we compute the cosine similarity matrix $S(l)$ between parallel sentence representations. MEXA measures the fraction of sentences whose true translation has the highest similarity in both directions:

\begin{equation}
\mu_{s,t}(l)
= \frac{1}{n}\sum_{i=1}^{n}
\mathbf{1}\!\left(S_{ii}(l) >
\max_{j \neq i} \{S_{ij}(l), S_{ji}(l)\}\right).
\label{eq:mexa}
\end{equation}

Following \citet{kargaran-etal-2025-mexa}, we use position-weighted sentence representations and evaluate alignment across layers, training checkpoints and all four tokenizer conditions.


\begin{figure}[h]
    \centering
    \includegraphics[width=\linewidth]{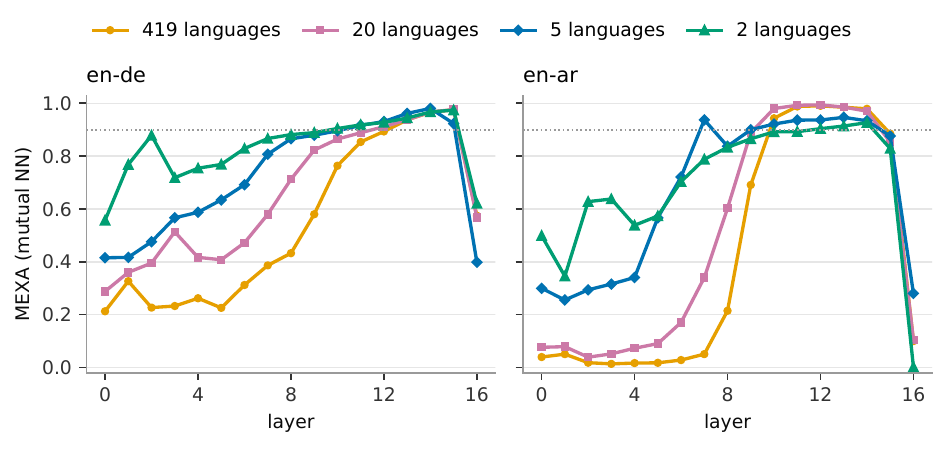}
    \caption{Representation alignment of en-de and en-ar 30B final checkpoint bilinguals across all four tokenizers trained on 419 (starved), 20, 5 (fair) and 2 languages.}
    \label{fig:mexa-all-four}
\end{figure}

\section{Results}
\label{sec:results}

\subsection{Representation Alignment}
\label{sec:results-alignment}

Figure~\ref{fig:mexa-3d} provides an overview of cross-lingual representation alignment across network layers and training checkpoints for the starved (419 languages) and fair (5 languages) tokenizers, while Figure~\ref{fig:mexa-all-four} shows final-checkpoint layer profiles for all four 64k tokenizers. Both figures show that tokenizers trained on less languages yield \emph{substantially higher} representation alignment in early-to-mid layers. The full trajectories in Figure~\ref{fig:mexa-3d} show that the fair tokenizer's early-to-mid-layer advantage appears by 2-5B training tokens and persists to 30B and alignment generally rises over training, especially for the cross-script bilinguals.

Across depth, alignment increases towards the later layers before dropping at the final layer, consistent with the language transition near the output reported in prior work \cite{Schut2025DoML, wendler-etal-2024-llamas, wang-etal-2025-lost-multilinguality}. Crucially, this advantage is mainly concentrated in layers 0-8, whereas alignment in layers 9-15 is already high under both tokenizers. Tokenization also changes \emph{where} high alignment emerges in the network. For \textsc{en--de}, the first layer reaching MEXA $\geq 0.90$ moves from layer 13 with the 419-language tokenizer to layers 12, 11, and 10 with the 20-, five-, and two-language tokenizers, respectively. Examining only the output layer can therefore miss the substantial effect of vocabulary allocation on sentence retrieval, helping reconcile our findings with \citet{limisiewicz-etal-2023-tokenization}.


\begin{figure}[t]
    \centering
    \includegraphics[width=\linewidth]{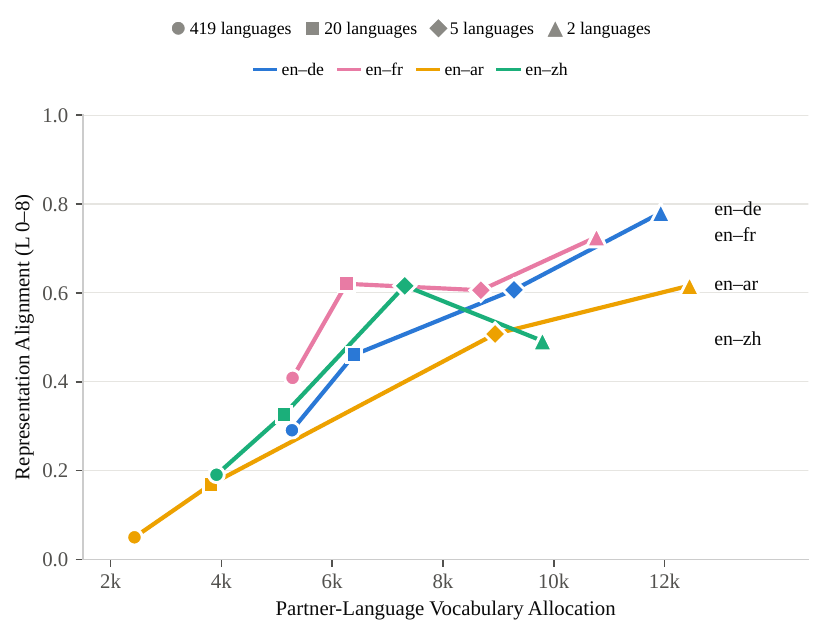}
    \caption{Mean MEXA representation alignment over layers 0-8 as a function of partner-language vocabulary allocation at 30B training tokens.}
    \label{fig:allocation-alignment}
\end{figure}

\paragraph{Alignment increases systematically with vocabulary allocation.}
To determine whether this finding reflects a graded relationship rather than a single tokenizer contrast, we analyze the relationship between vocabulary allocation and alignment across all four tokenizer conditions. Figure~\ref{fig:allocation-alignment} shows a strong positive relationship between allocation and early-to-mid-layer alignment. At 30B training tokens, alignment generally increases from 419 to 20 to five languages, with further gains under pair-specific tokenizers for German, French, and Arabic. However, the trend is not strictly monotonic. French changes little between 20 and five languages, while Chinese alignment decreases from five to two. One possible explanation is that a relatively small set of characters covers most written Chinese, so additional vocabulary increasingly produces multi-character tokens with diminishing benefits for alignment. Alphabetic languages, in contrast, may benefit from a larger vocabulary of meaningful subword units. Overall, alignment increases with vocabulary allocation, though not at every step.

\begin{figure}[h]
\centering
\includegraphics[width=\linewidth]{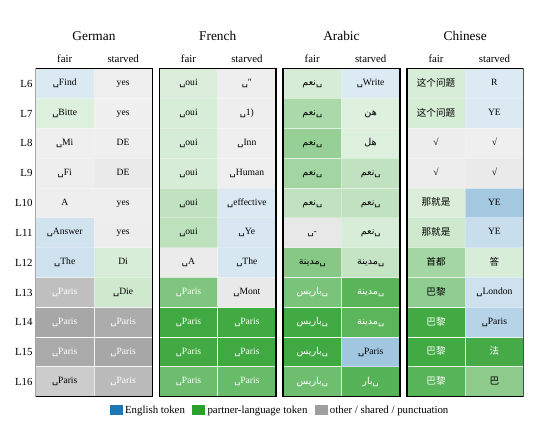}
\caption{The logit lens applied to bilingual models on the first token continuation of the input ``\emph{The capital of France is:}'', translated into each partner language.}
\label{fig:logit-lens-qualitative}
\end{figure}

\begin{figure}[t]
  \centering
  \includegraphics[width=0.8\linewidth]{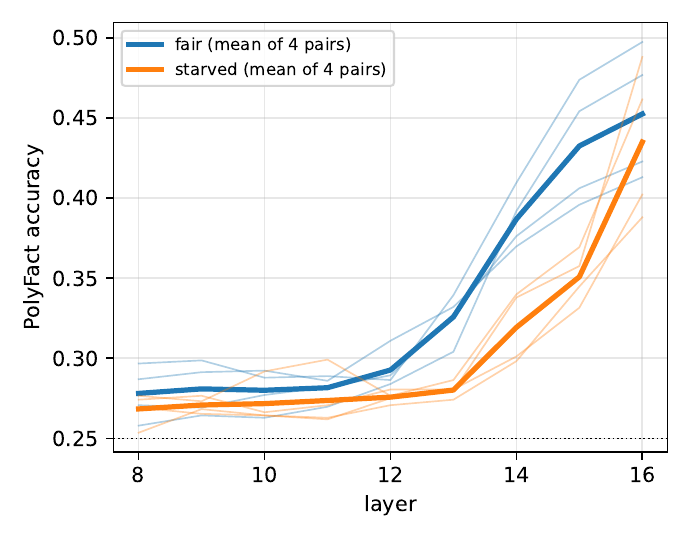}
  \caption{PolyFact accuracy by layer under the logit lens for the 30B bilingual models. Curves for all four tokenizer conditions are provided in Appendix \ref{app:alignment}, Figure}
  \label{fig:lens-interior}
\end{figure}

\paragraph{Higher alignment might stem from earlier decoding.}
One possible explanation for higher representation alignment is that better vocabulary allocation makes factual knowledge decodable earlier in the network. Applying the logit lens (Figure~\ref{fig:logit-lens-qualitative}), we observe that while early layers produce unrelated tokens such as ``Human'' or ``Yes'' (ar), the correct answer (e.g., ``Paris'') emerges earlier under the fair tokenizer across all four languages. We validate this quantitatively in Figure~\ref{fig:lens-interior}, showing that factual recall accuracy rises at earlier layers under fair tokenization. Together, these findings suggest that improved vocabulary allocation shifts the emergence of decodable factual representations towards earlier layers, potentially explaining the observed increase in early-to-mid-layer alignment.

\subsection{Cross-lingual Transfer}
\label{sec:results-transfer}

Having established that fairer vocabulary allocation strengthens representation alignment, we next ask whether it also increases the benefits of bilingual pretraining. We measure bilingual transfer by comparing each bilingual model against a monolingual baseline at matched target-language token exposure (Equation~\ref{eq:bts-loss}), i.e. measuring how much higher (or lower) validation loss of a bilingual model is compared to its monolingual counterpart. To ensure comparability under same learning rate, curves are reported until 11B target-language tokens, since that equals 22B total bilingual tokens, marking the end of the stable phase of WSD.


\begin{figure}[t]
    \centering
    \includegraphics[width=\linewidth]{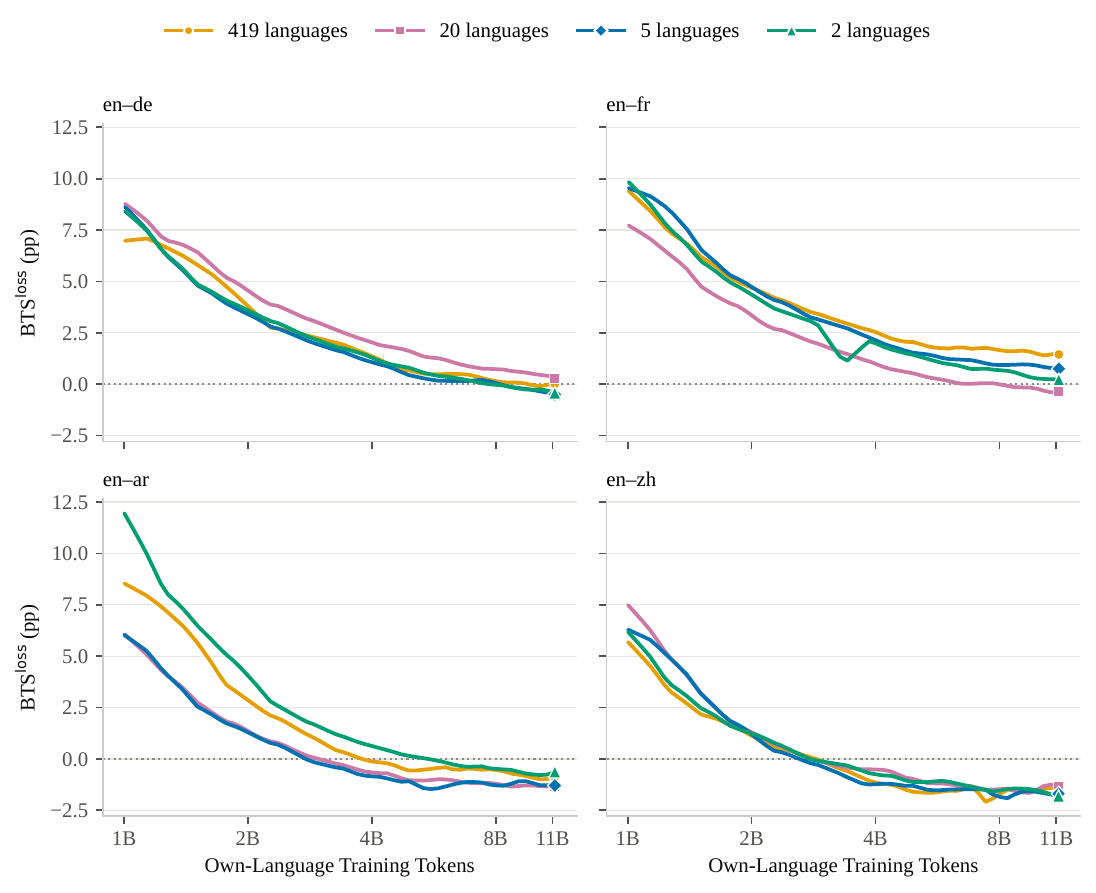}
    \caption{Bilingual transfer into the partner language $\mathrm{BTS}^{\mathrm{loss}}$ at matched target-language exposure under all four tokenizers over training. For transfer into English refer to Appendix \ref{app:bts}, Figure \ref{fig:bts-english}.}
    \label{fig:bts-absolute}
\end{figure}

\paragraph{Transfer depends on the language pair and direction, not the tokenizer.}
Figure~\ref{fig:bts-absolute} shows bilingual transfer into the partner language under all four tokenizers over training. Over training tokens, bilingual transfer generally decreases and converges at around 8-10B tokens.  When English is added to the partner language's training data, French benefits most, German is approximately neutral, and Arabic and Chinese show negative transfer. Crucially, this ordering persists under all four tokenizers. Among the four pairs studied, the two same-script pairs (en-de, en-fr) consistently exhibit higher transfer into the partner language than the two cross-script pairs (en-ar, en-zh), reinforcing the idea of a script barrier. 

\TokenizerEffects

Table \ref{tab:tok-effects} isolates the effect of the tokenizer by expressing each tokenizer's transfer as its deviation from the average of all four tokenizers, averaged over language pairs. All effects lie within half a percentage point of this average, and only one confidence interval (20-lang into English) excludes zero, which suggests that tokenizer choice has at most a marginal influence on cross-lingual transfer. When measuring transfer into English, bilingual training generally produces negative transfer for all languages including French and German. This is consistent with the observation from prior work that English can be a strong donor but a weak recipient \cite{protasov-etal-2024-super}.

Together, these results reveal an \emph{alignment-transfer dissociation}. While better vocabulary allocation substantially improves early-to-mid-layer representation alignment, it has little effect on bilingual transfer, suggesting that language-pair characteristics matter more than tokenizer quality. This also strengthens the conclusions of \citet{longpre2026atlas} and \citet{chang-etal-2024-multilinguality}, whose findings relied on tokenizers with substantial vocabulary imbalances across scripts. Our results suggest that their transfer findings are unlikely to be artifacts of these imbalances and that stronger representation alignment does not necessarily translate into better cross-lingual transfer.


\subsection{Parametric Knowledge Sharing}
\label{sec:results-sharing}

The alignment-transfer dissociation raises the question of whether improved vocabulary allocation also strengthens cross-lingual factual knowledge sharing. We examine facts answered correctly in both languages with distinct answer surface forms (leaving out answers like ``Paris'' which are identical across same-script languages) and quantify parametric sharing using the causal ablation transfer rate (CTR; Equation~\ref{eq:causal-transfer-rate}, Figure~\ref{fig:causal-sharing}), which measures how much ablation damage transfers across languages, normalized by a other-fact baseline. 

\begin{figure}[t]
\centering
\includegraphics[width=\linewidth]{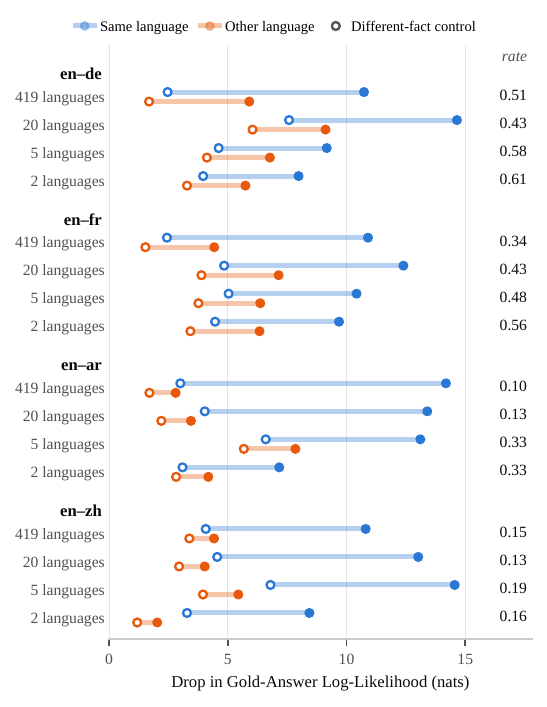}
\caption{Causal ablation test for the 30B bilingual models, on PolyFact facts (461-697 facts per model). CTR is the ratio of the orange line to the blue line. Zeroing 100 random neurons costs $\leq 0.02$ nats.}
\label{fig:kn-ablation-4tok}
\end{figure}

\paragraph{Causal evidence of shared factual knowledge.}
Figure~\ref{fig:kn-ablation-4tok} shows that ablating neurons identified in another language consistently reduces factual recall beyond the different-fact baseline, providing causal evidence of cross-lingual knowledge sharing. However, same-language ablations generally cause substantially greater damage, indicating that factual knowledge is only partially shared. Interestingly, the script barrier is also evident in parametric sharing, where the causal transfer rate (CTR) is consistently higher for the same-script pairs (\textsc{en--de}, \textsc{en--fr}) than for the cross-script pairs (\textsc{en--ar}, \textsc{en--zh}). Furthermore, the illustration highlights the importance of a mismatched-fact baseline, as the different-fact control ablation demonstrates substantially higher damage to the gold-answer log-likelihood (3-7 nats) compared to random neurons ($\leq 0.02$). Notably, higher CTR does not necessarily imply greater absolute cross-lingual ablation damage. For German, both cross-language and same-language ablation effects decrease with fewer tokenizer languages (i.e. higher vocab allocation), but the latter decreases more strongly, resulting in a higher CTR (0.51 to 0.61).

\begin{figure}[t]
    \centering
    \includegraphics[width=\linewidth]{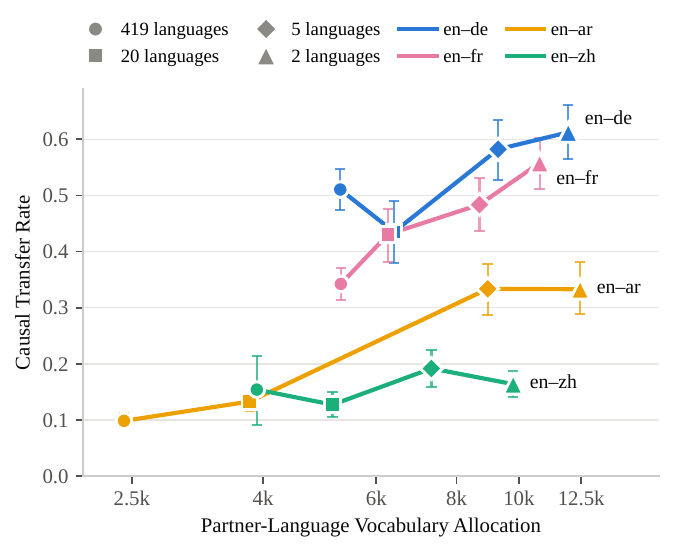}
    \caption{Parametric knowledge sharing against partner-language vocabulary allocation (30B finals) with 95\% CIs. }
    \label{fig:allocation-sharing}
\end{figure}

\paragraph{Vocabulary allocation strengthens sharing, but unevenly.}
Figure~\ref{fig:allocation-sharing} shows that parametric sharing measured by CTR generally increases with vocabulary allocation for German, French, and Arabic, although not strictly monotonically. Most notably, CTR rises from $0.34$ to $0.56$ for French and from $0.10$ to $0.33$ for Arabic between the 419- and 2-language tokenizers. Chinese, however, shows little systematic improvement, consistent with our representation alignment results and our hypothesis of diminishing returns for multi-character tokens. Thus, unlike bilingual transfer, parametric knowledge sharing is sensitive to vocabulary allocation, although its magnitude remains strongly dependent on the language pair. Additional correlational analyses support the causal results by measuring knowledge-neuron overlap, provided in Appendix~\ref{app:neuron-overlap}, Figure \ref{fig:knowledge-overlap}.

\subsection{Controlling for Language-Modeling Quality}
\label{sec:results-controls}

\begin{figure}[t]
  \centering
  \includegraphics[width=\linewidth]{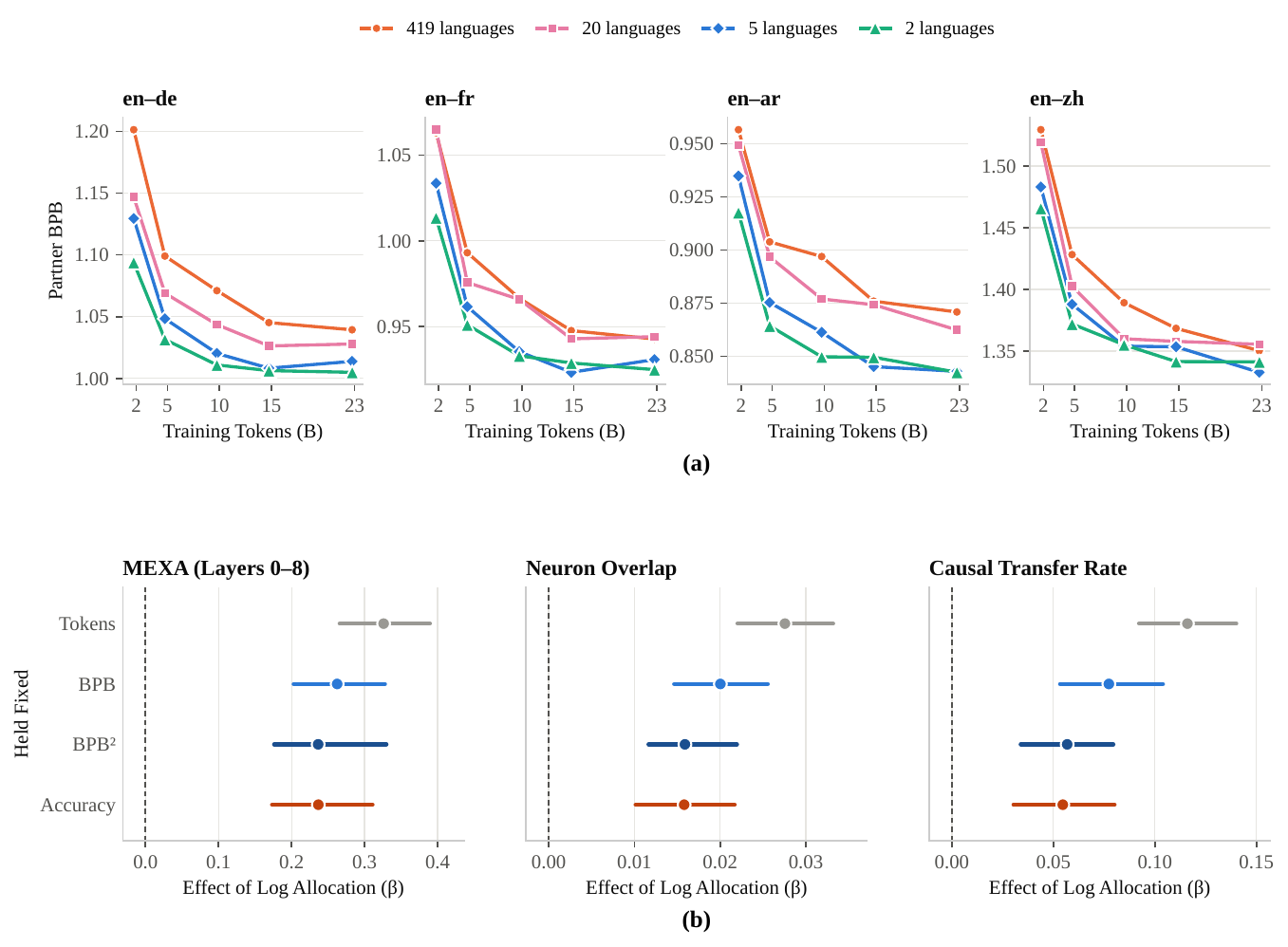}
  \caption{Controlling for model quality. (a) Tokenizers covering fewer languages reach a lower partner-language validation loss (BPB) throughout training. (b) Effect of vocabulary allocation on alignment and knowledge sharing, holding fixed training tokens, BPB (linear or quadratic: BPB$^2$) or factual accuracy (95\% CIs).}
  \label{fig:matched-loss}
\end{figure}

Tokenizers trained on fewer languages produce shorter sequences, so the model reads more text per token and reaches a lower validation loss (BPB) in every language pair (Figure~\ref{fig:matched-loss}a). This raises the question: Are the gains in alignment and sharing then just a by-product of a better model? To test this, we ask whether more vocabulary allocation still helps among checkpoints of \emph{equal quality}. 

Since exactly matched checkpoints rarely exist, we fit a linear model to all 80 stable-phase checkpoints (2--23B tokens, four tokenizers, four pairs) that predicts each metric from log allocation and a quality measure at the same time, with a separate baseline per language pair. The allocation coefficient $\beta$ then gives the effect of allocation at equal quality, and $\beta = 0$ would mean no effect beyond quality. As quality measures we use, in turn, training tokens (the reference, with no quality correction), partner-language BPB, BPB with a curved, pair-specific fit (BPB$^2$), and partner-language factual accuracy. The 95\% confidence intervals come from bootstrapping checkpoints within pairs.

The effect stays positive under every control with all intervals excluding zero (Figure~\ref{fig:matched-loss}b). For alignment, about three quarters of the effect remains ($\beta = 0.33 \to 0.24$-$0.26$). For sharing, about half to two thirds remains, both for neuron overlap ($0.028 \to 0.016$-$0.020$) and for the causal transfer rate ($0.116 \to 0.055$-$0.077$). Model quality thus explains part of the sharing gain but little of the alignment gain. In fact, these estimates may be conservative, since better-tokenized models reach the same validation loss after fewer training tokens and optimization steps, potentially leaving less opportunity for alignment and knowledge sharing to develop. We treat this as a robustness check rather than proof.

\section{Conclusion}
\label{sec:conclusion}

Our results reveal two novel effects of improved vocabulary allocation beyond raising each language's individual performance. First, improved tokenization yields substantially higher cross-lingual representation alignment in early-to-mid layers, accompanied by a depth-of-emergence effect in which models commit to their output tokens at shallower layers. Second, and perhaps most consequentially, it leads to greater cross-lingual parametric knowledge sharing in FFN neurons, supported by both correlational and causal evidence. Facts increasingly co-locate in the same parameters, pointing to improved allocation as a potential lever on the parameter competition underlying the curse of multilinguality. Despite substantially improving representation alignment and knowledge sharing, improved tokenization does not increase cross-lingual transfer, revealing an \emph{alignment-transfer dissociation} and leaving the same-script cross-script transfer gap unchanged, demonstrating that the script barrier is not a tokenizer artifact. We hope our work motivates future research to adopt more equitable experimental setups when studying cross-lingual interactions.

\section*{Limitations}

While our design isolates the tokenizer as the only experimental variable, several limitations remain. First, with only two language pairs per script group, script is confounded with language identity. Our results establish that the transfer gap is not a tokenizer artifact, but its interpretation as a script effect relies on prior large-scale evidence disentangling script from language family \cite{longpre2026atlas}. We therefore treat the group-level gaps as descriptive. Disentangling script from language identity would require a broader set of languages, which is beyond our computational budget, since studying transfer across our five languages and four tokenizer conditions already required 39 training runs.

Second, all findings are established at a single scale of 1B parameters, a 30B-token training budget, and one seed per configuration. While this is the smallest scale at which prior data and tokenizer ablations report reliable signal, whether the alignment-transfer dissociation persists at larger scales, longer training horizons, or across seeds remains open. Nevertheless, our central patterns replicate across all four language pairs and all four tokenizer conditions, with each configuration corresponding to an independent training run, so the findings do not depend on any single model.

Finally, our models are bilingual with English as a fixed anchor. While this reflects common practice, it leaves transfer between two non-English languages and genuinely multilingual settings unexplored, particularly settings with many training languages where the curse of multilinguality may be more pronounced. In addition, we restrict our study to high-resource languages to approximately control for data quality. 

\section*{Acknowledgements}

This work was supported by the Engineering and Physical Sciences Research Council (EPSRC) [grant number UKRI5028], a 2025 Amazon Research Award, and GPU computing resources provided through the NVIDIA Academic Grant Program.

\bibliography{references}

\appendix

\section{Experimental Details}
\label{app:setup}

\paragraph{Training Data.}
Table~\ref{tab:data-stats} summarizes the training corpora. All languages use FineWeb2-HQ \cite{messmer-etal-2025-enhancing}, except English, which uses FineWeb-HQ. Arabic has only 94\,GB of available data and is therefore repeated for 1.67 epochs under the starved tokenizer at 30B tokens. Prior work finds negligible degradation from repeating data for up to four epochs \cite{scaling2023muennighoff}.

\begin{table}[h]
\centering
\small
\setlength{\tabcolsep}{4pt}
\begin{tabular}{lrrr}
\toprule
Language & Size (GB) & Epochs & Byte premium \\
\midrule
English & 14{,}751 & 1 & 1.000 \\
German  & 618 & 1 & 1.186 \\
French  & 483 & 1 & 1.237 \\
Arabic  & 94 & 1.67 & 1.600 \\
Chinese & 784 & 1 & 0.925 \\
\bottomrule
\end{tabular}
\caption{Training corpus sizes, epochs under the starved tokenizer at 30B tokens, and byte-premium ratios relative to English, measured on FLORES+.}
\label{tab:data-stats}
\end{table}

\paragraph{Tokenizer Training.}
Our starved tokenizer uses the same 419-language coverage and sampling temperature ($T=100$) as \citet{longpre2026atlas}, but is trained on FineWeb2 rather than MADLAD-400 to minimize differences in tokenizer training data. The fair tokenizer uses byte-premium-adjusted sampling, with ratios measured on FLORES+ \cite{gordeev-etal-2024-flores}, to approximately equalize semantic content across languages.

\paragraph{Starved Tokenizer Replication.}
\label{app:tokenizer-replication}
After training our models, we obtained the original tokenizer from \citet{longpre2026atlas} and compared both implementations. They share the same Unigram algorithm, byte fallback, character coverage ($0.999995$), and sampling temperature ($T=100$). Despite different training corpora, their vocabularies overlap by 71\%, with fertility differences of only 1--3 tokens per sentence for English, German, French, and Arabic, and identical segmentation on our Chinese test sentences.

\paragraph{Architecture and Hyperparameters.}
All models use full multi-head attention, untied input/output embeddings, and SwiGLU feed-forward layers of size 5632. Table~\ref{tab:hyperparams} reports the training configuration. We use warmup-stable-decay (WSD) rather than cosine decay to permit comparisons between checkpoints at identical learning rates during the stable phase.

\begin{table}[h]
\centering
\small
\setlength{\tabcolsep}{4pt}
\begin{tabular}{ll}
\toprule
Hyperparameter & Value \\
\midrule
Parameters & 1.09B \\
Transformer layers & 16 \\
Hidden / FFN size & 2048 / 5632 \\
Attention heads & 16 \\
Vocabulary size & 65{,}536 \\
Context length & 2048 \\
Optimizer & AdamW \\
$\beta_1,\beta_2,\epsilon$ & $0.9,\,0.95,\,10^{-8}$ \\
Weight decay & 0.1 \\
Gradient clipping & 1.0 \\
Peak learning rate & $3\times10^{-3}$ \\
Minimum learning rate & $3\times10^{-4}$ \\
LR schedule & WSD (1B / 23B / 6B) \\
Batch size & 1M tokens \\
Precision & bf16 \\
Training budget & 30B tokens \\
\bottomrule
\end{tabular}
\caption{Model architecture and training hyperparameters.}
\label{tab:hyperparams}
\end{table}

\paragraph{Model List}
\label{app:model-list}

\ModelList

\section{Training and Downstream Performance}
\label{app:training}

\begin{figure*}[t]
\centering
\includegraphics[width=\textwidth]{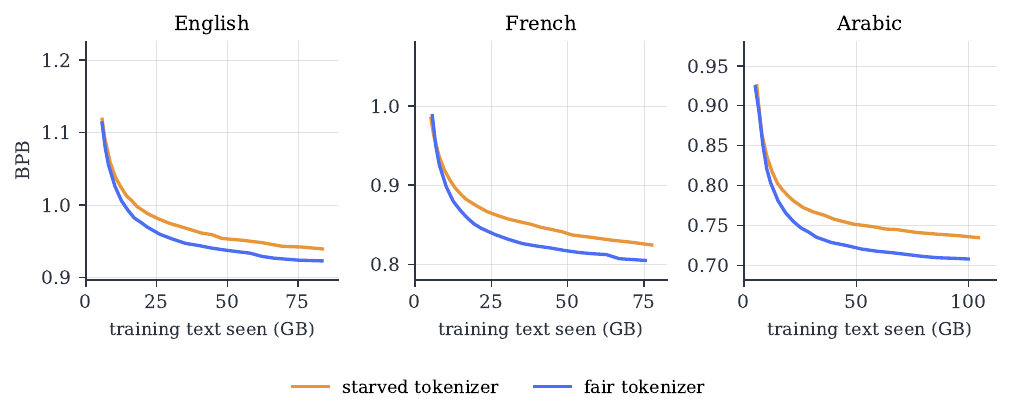}
\caption{Monolingual validation BPB at equal amounts of training text.}
\label{fig:sample-efficiency}
\end{figure*}

\paragraph{Sample Efficiency.}
At equal token budgets, better compression allows the fair tokenizer to process 20--48\% more text, depending on the language. To assess whether this alone explains its lower validation loss, Figure~\ref{fig:sample-efficiency} compares models at equal amounts of training text. The advantage decreases but persists: for English, the BPB difference shrinks from $-0.021$ to $-0.017$, and for Arabic from $-0.034$ to $-0.028$. Thus, improved compression alone does not fully explain the language-modeling gains.

\begin{figure}[t]
\centering
\includegraphics[width=\linewidth]{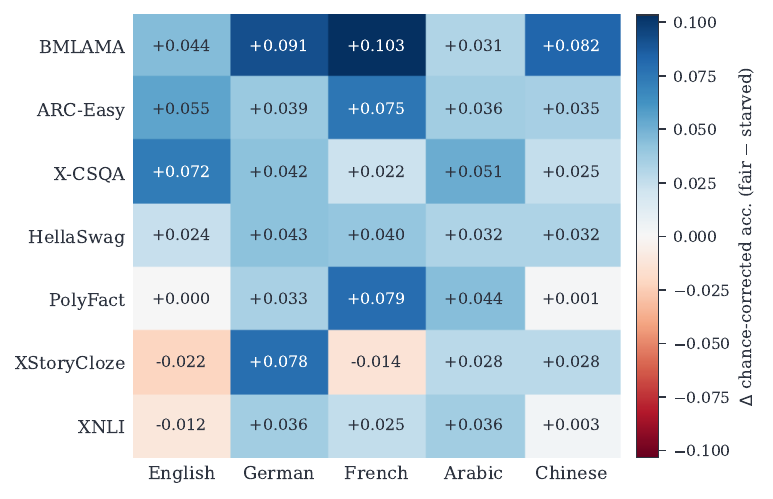}
\caption{Change in chance-corrected benchmark accuracy from the starved to the fair tokenizer.}
\label{fig:bench-heatmap}
\end{figure}

\paragraph{Downstream Performance.}
We evaluate seven multilingual benchmarks using cloze-style multiple-choice prompts scored by candidate log-likelihood. Figure~\ref{fig:bench-heatmap} shows that fair tokenization generally improves downstream accuracy, with BMLAMA factual recall benefiting most ($+7$ percentage points on average). Across languages, German benefits most ($+5.2$ points on average), while English benefits least ($+2.3$ points).

\section{Additional Bilingual Transfer Results}
\label{app:bts}

\begin{figure}[t]
\centering
\includegraphics[width=\linewidth]{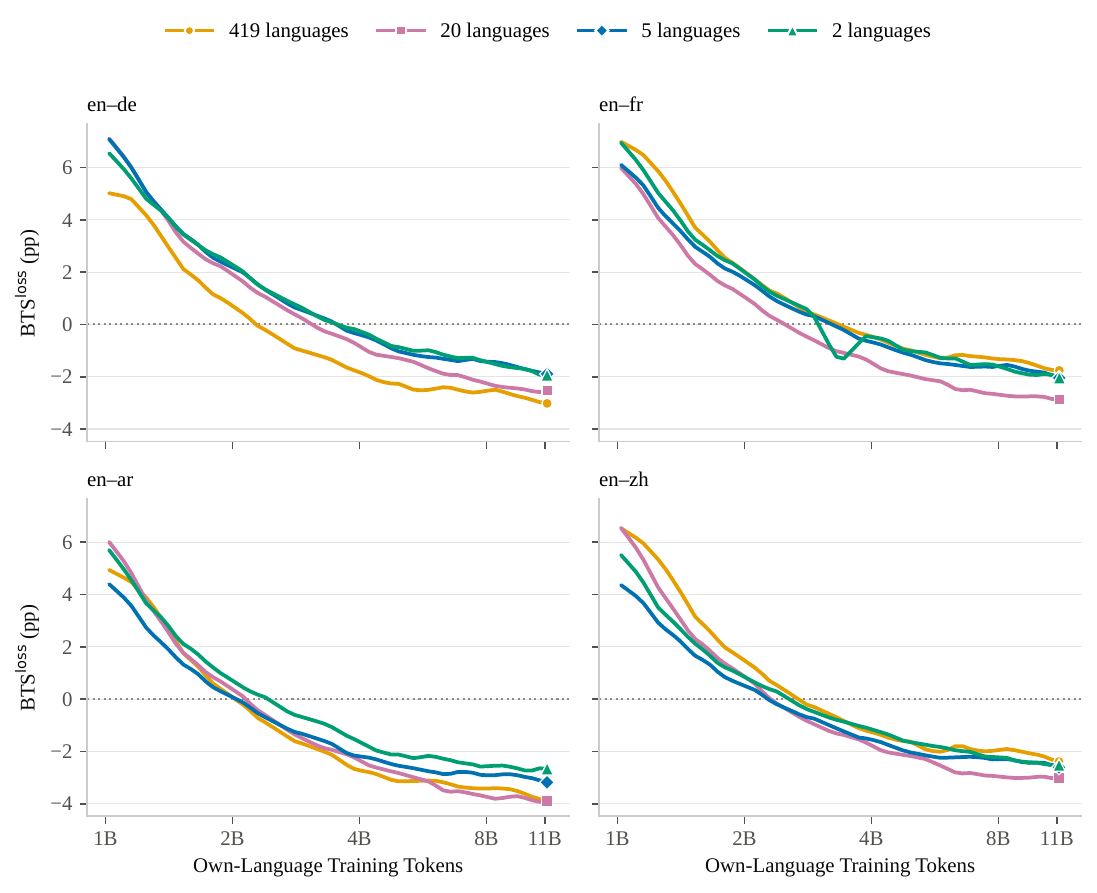}
\caption{Bilingual transfer into English under all four tokenizer conditions at matched target-language exposure.}
\label{fig:bts-english}
\end{figure}

\paragraph{Transfer into English.}
Figure~\ref{fig:bts-english} shows bilingual transfer into English at matched target-language exposure. Unlike transfer into partner languages, transfer into English is consistently negative for all four pairs and becomes more negative over training. This asymmetry is consistent with English acting as a stronger donor than recipient \cite{protasov-etal-2024-super}. Importantly, tokenizer allocation does not substantially alter this pattern.

\paragraph{Downstream Performance Transfer.}
We additionally measure transfer using downstream accuracy. Since benchmarks have different chance levels $c_b$, we first normalize accuracy as $A'_b=(A_b-c_b)/(1-c_b)$ and average across benchmarks. Transfer is then defined as

\begin{equation}
\mathrm{BTS}^{\mathrm{perf}}_{s\to t}(d)
=\frac{\bar{A}'^{\,\mathrm{bi}}_t(d)-
       \bar{A}'^{\,\mathrm{mono}}_t(d)}
      {1-\bar{A}'^{\,\mathrm{mono}}_t(d)}.
\label{eq:bts-perf}
\end{equation}

\begin{figure*}[t]
\centering
\includegraphics[width=\textwidth]{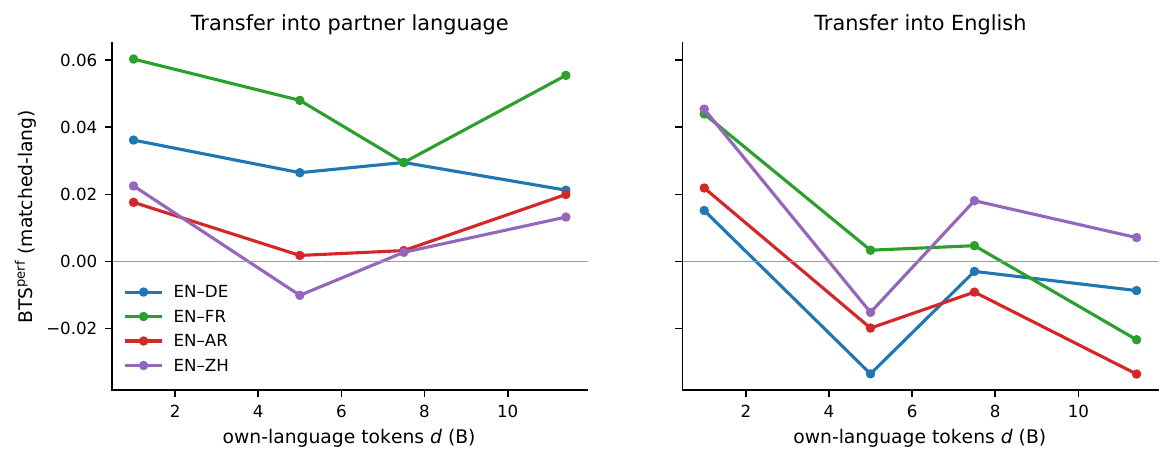}
\caption{Bilingual transfer measured through downstream performance under the starved tokenizer.}
\label{fig:bts-perf}
\end{figure*}

Figure~\ref{fig:bts-perf} shows that downstream transfer is more checkpoint-dependent than loss-based transfer. French generally benefits most, while the relative ordering of other languages varies over training. We therefore use validation-loss transfer as our primary measure.

\section{Additional Cross-lingual Alignment Analyses}
\label{app:alignment}

\begin{figure*}[t]
\centering
\includegraphics[width=0.8\textwidth]{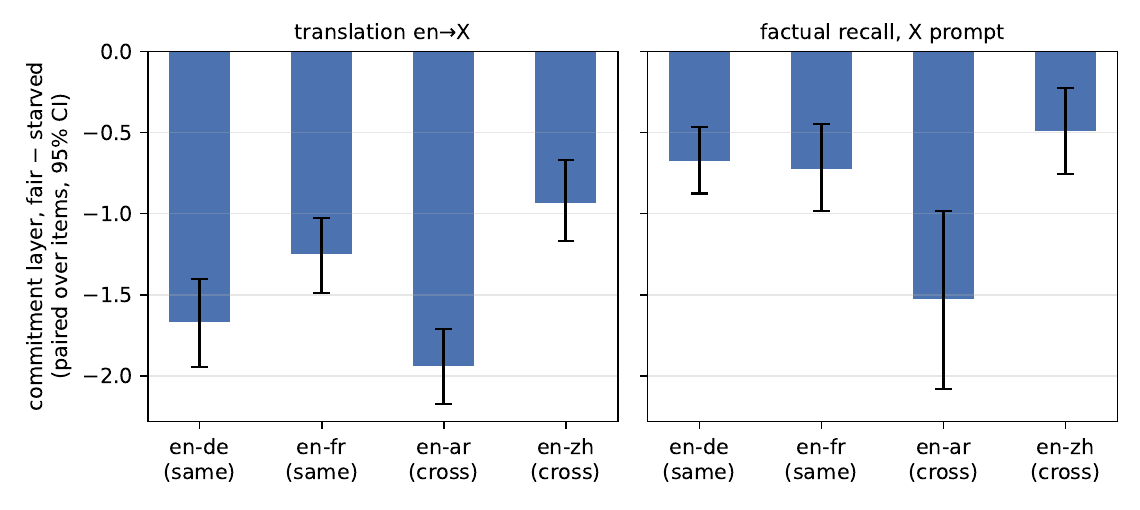}
\caption{Difference in output-token commitment layer between fair and starved models (95\% CIs), measured on translation and PolyFact factual recall.}
\label{fig:commitment-layer-deltas}
\end{figure*}

\begin{figure*}[t]
\centering
\includegraphics[width=\textwidth]{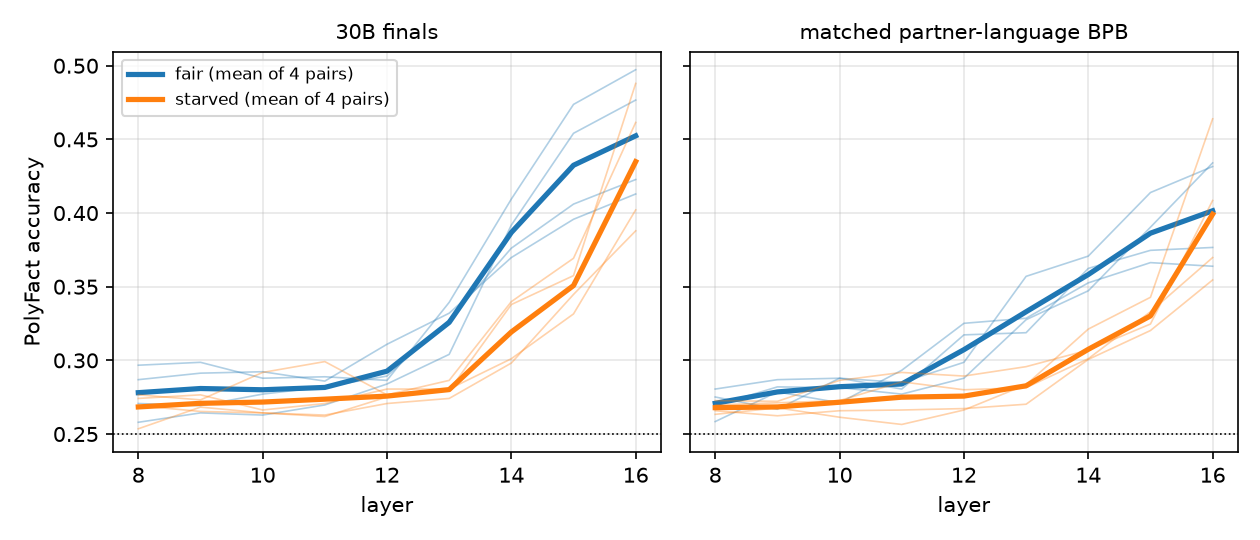}
\caption{PolyFact accuracy across layers, averaged over four languages. Factual answers become decodable earlier under the fair tokenizer.}
\label{fig:logit-lens-polyfact}
\end{figure*}

\paragraph{Earlier Emergence of Decodable Representations.}
We quantify the depth-of-emergence effect using the logit lens, recording the first layer at which the projected prediction matches the final output token. Figure~\ref{fig:commitment-layer-deltas} shows that the fair tokenizer generally leads to earlier commitment across languages, with reductions of approximately $0.2$--$2$ layers. The effect is particularly pronounced for Arabic translation. Figure~\ref{fig:logit-lens-polyfact} further shows that factual recall becomes decodable at earlier layers, consistent with the earlier emergence of high representation alignment.

\begin{figure}[t]
  \centering
  \includegraphics[width=\linewidth]{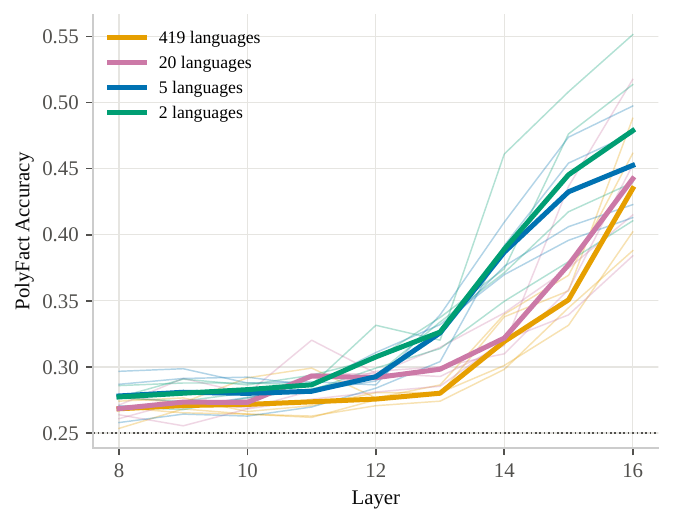}
  \caption{PolyFact accuracy by layer under the logit lens for the 30B bilingual models of all four tokenizers, with prompts in the partner language. Thin lines show individual language pairs and thick lines their mean; the dotted line marks chance (0.25).}
  \label{fig:lens-interior-4tok}
\end{figure}

\begin{figure}[t]
\centering
\includegraphics[width=\linewidth]{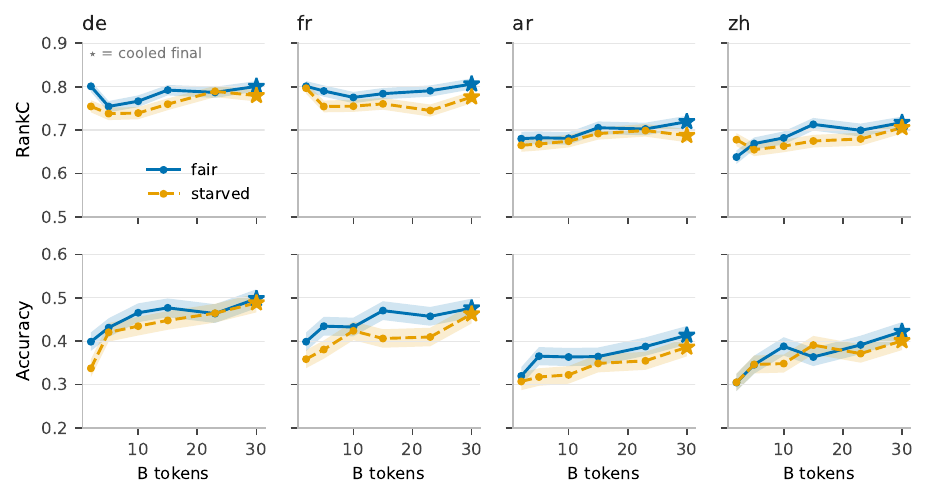}
\caption{Cross-lingual consistency and factual accuracy on PolyFact across training checkpoints.}
\label{fig:rankc_traj}
\end{figure}

\paragraph{Cross-lingual Consistency.}
Figure~\ref{fig:rankc_traj} compares cross-lingual consistency (RankC) and factual accuracy over training. While accuracy improves substantially, consistency is largely established early: same-script pairs reach approximately $0.75$--$0.80$ after 2B tokens and change little thereafter. Cross-script pairs improve by only $0.04$--$0.07$. This contrasts with representation alignment, which continues developing throughout pretraining.

\section{Parametric Knowledge Sharing: Robustness}
\label{app:neuron-overlap}

\begin{figure}[t]
\centering
\includegraphics[width=\linewidth]{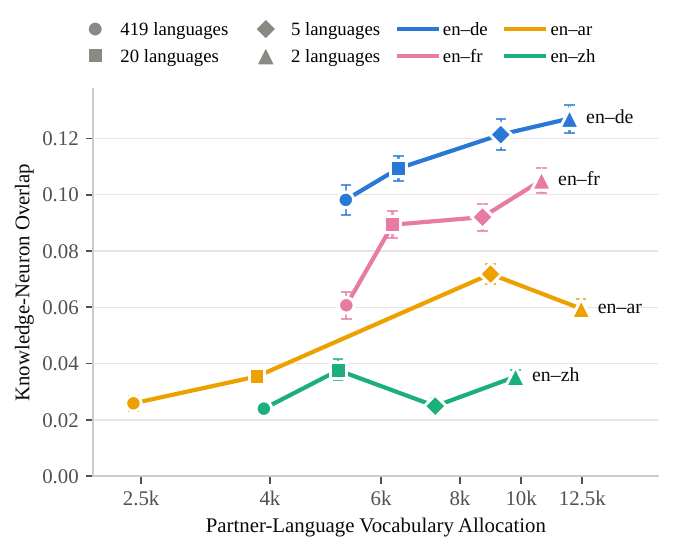}
\caption{Fact-specific knowledge-neuron overlap against partner-language vocabulary allocation at 30B tokens.}
\label{fig:knowledge-overlap}
\end{figure}

\paragraph{Knowledge-Neuron Overlap.}
Figure~\ref{fig:knowledge-overlap} complements the causal transfer analysis with cross-lingual Jaccard overlap, subtracting the expected overlap between neurons identified for different facts. Consistent with the causal results, vocabulary allocation generally increases fact-specific neuron overlap for German, French, and Arabic, while Chinese exhibits a considerably weaker effect.

\paragraph{Robustness to Neuron Cutoff.}
Table~\ref{tab:dks-robustness} shows that the fair tokenizer's advantage in fact-specific neuron overlap persists across $K\in\{50,100,200,500\}$. German, French, and Arabic consistently improve, while Chinese shows only small differences.

\begin{table}[t]
\centering
\small
\setlength{\tabcolsep}{3pt}
\renewcommand{\arraystretch}{1.15}
\begin{tabular}{lcccc}
\toprule
Pair & $K=50$ & $K=100$ & $K=200$ & $K=500$ \\
\midrule
en--de
& +.021 & +.025 & +.027 & +.031 \\
& {\scriptsize [+.013,+.030]}
& {\scriptsize [+.018,+.032]}
& {\scriptsize [+.021,+.033]}
& {\scriptsize [+.026,+.036]} \\
\addlinespace[3pt]
en--fr
& +.028 & +.031 & +.030 & +.033 \\
& {\scriptsize [+.021,+.036]}
& {\scriptsize [+.026,+.037]}
& {\scriptsize [+.025,+.035]}
& {\scriptsize [+.029,+.038]} \\
\addlinespace[3pt]
en--ar
& +.056 & +.047 & +.042 & +.038 \\
& {\scriptsize [+.051,+.062]}
& {\scriptsize [+.043,+.051]}
& {\scriptsize [+.039,+.046]}
& {\scriptsize [+.035,+.041]} \\
\addlinespace[3pt]
en--zh
& +.001 & +.001 & +.003 & +.004 \\
& {\scriptsize [-.004,+.007]}
& {\scriptsize [-.002,+.005]}
& {\scriptsize [-.000,+.005]}
& {\scriptsize [+.002,+.006]} \\
\bottomrule
\end{tabular}
\caption{Robustness of fact-specific knowledge-neuron overlap
to the top-$K$ cutoff. Values show Fair $-$ Starved
differences in overlap, with 95\% confidence intervals
beneath.}
\label{tab:dks-robustness}
\end{table}

\paragraph{Controlling for Factual Recall.}
To assess whether increased sharing merely reflects higher factual accuracy, we compare checkpoints matched within one percentage point of partner-language PolyFact accuracy. Table~\ref{tab:sharing-matched-recall} shows that the fair tokenizer generally retains higher neuron overlap and causal transfer rates under this control, with Chinese again showing the weakest response.

\begin{table}[t]
\centering
\small
\setlength{\tabcolsep}{4pt}
\begin{tabular}{llrr}
\toprule
Partner & Matched checkpoints & $\Delta$ overlap & $\Delta$ CTR \\
\midrule
de & 23B vs.\ 23B & $+.032^{*}$ & $+.056^{*}$ \\
fr & 10B vs.\ 10B & $+.007$ & $+.054^{*}$ \\
ar & 15B vs.\ 23B & $+.036^{*}$ & $+.120^{*}$ \\
zh & 10B vs.\ 15B & $+.008^{*}$ & $-.005$ \\
\bottomrule
\end{tabular}
\caption{Parametric sharing at matched factual recall (fair $-$ starved). Checkpoints differ by at most 1 percentage point in partner-language PolyFact accuracy and lie within the stable training phase. $^{*}$ denotes a 95\% bootstrap CI excluding zero.}
\label{tab:sharing-matched-recall}
\end{table}

\section{Source Code}
Source code and model checkpoints are available at
\url{https://anonymous.4open.science/r/XScript-Pretraining-113E}.

\end{document}